\documentclass[runningheads]{llncs}

\usepackage{eccv}

\usepackage{eccvabbrv}

\usepackage{graphicx}
\usepackage{booktabs}
\usepackage{tikz}
\usetikzlibrary{positioning,shapes.geometric, backgrounds, fit}
\usepackage{pgfplots}
\pgfplotsset{compat=1.18}
\usepackage{bm}
\usepackage{xcolor}
\usepackage{tabu}

\newcommand{\R}{\mathbb{R}}

\usepackage[accsupp]{axessibility}  
\usepackage[mode=buildnew]{standalone}

\newcommand{\cotronlvsapce}{\vspace{-0.0cm}}

\usepackage{multirow}

\definecolor{darkspringgreen}{rgb}{0.09, 0.45, 0.27}

\newcommand{\AJ}[1]{{\textcolor{cyan}{[\textbf{AJ} #1]}}}
\newcommand{\AK}[1]{{\textcolor{red}{[\textbf{AK} #1]}}}
\newcommand{\AW}[1]{{\textcolor{green}{[\textbf{AW} #1]}}}

\definecolor{cvprblue}{rgb}{0.21,0.49,0.74}
\definecolor{cvprred}{rgb}{0.74,0.21,0.49}
\usepackage{hyperref}

\usepackage{orcidlink}

\begin{document}

\title{iNeMo: Incremental Neural Mesh Models for \\Robust Class-Incremental Learning} 

\titlerunning{iNeMo}

\author{Tom Fischer\inst{1} \orcidlink{0009-0009-6776-2767} \and
Yaoyao Liu\inst{2}\orcidlink{0000-0002-5316-3028} \and
Artur Jesslen\inst{3} \and
Noor Ahmed\inst{1}\orcidlink{0009-0002-0084-0141} 
\and\\
Prakhar Kaushik\inst{2}\orcidlink{0000-0001-6449-8088} \and
Angtian Wang\inst{2}\orcidlink{0009-0006-9189-5277} \and
Alan Yuille\inst{2} \and \\
Adam Kortylewski\inst{3,4} \and
Eddy Ilg\inst{1} }

\authorrunning{T.~Fischer et al.}

\institute{Saarland University, Saarbrücken, Germany\\
\email{\{fischer, ilg\}@cs.uni-saarland.de}
\and
Johns Hopkins University, Baltimore, USA\\
\email{\{yliu538, angtianwang, ayuille1\}@jhu.edu}
\and
University of Freiburg, Freiburg, Germany\\
\email{\{jesslen, kortylew\}@cs.uni-freiburg.de}
\and
Max-Planck Institute for Informatics, Saarbrücken, Germany\\
\email{akortyle@mpi-inf.mpg.de}}


\maketitle

\begin{abstract}
Different from human nature, it is still common practice today for vision tasks to train deep learning models only initially and on fixed datasets. A variety of approaches have recently addressed handling continual data streams. However, extending these methods to manage out-of-distribution (OOD) scenarios has not effectively been investigated. On the other hand, it has recently been shown that non-continual neural mesh models
exhibit strong performance in generalizing to such OOD scenarios. 
To leverage this decisive property in a continual learning setting, we propose incremental neural mesh models that can be extended with new meshes over time. In addition, we present 
a latent space initialization strategy that enables us to allocate feature space for future unseen classes in advance and a positional regularization term that forces the features of the different classes to consistently stay in respective latent space regions. We demonstrate the effectiveness of our method through extensive experiments on the Pascal3D and ObjectNet3D datasets and show that our approach outperforms the baselines for classification by $2-6\%$ in the in-domain and by $6-50\%$ in the OOD setting. Our work also presents the first incremental learning approach for pose estimation. 
Our code and model can be found at \href{https://github.com/Fischer-Tom/iNeMo}{github.com/Fischer-Tom/iNeMo}.

\end{abstract}


\section{Introduction}
\label{sec:intro}

\begin{figure}[t]
    \centering
    \includegraphics[width=\textwidth]{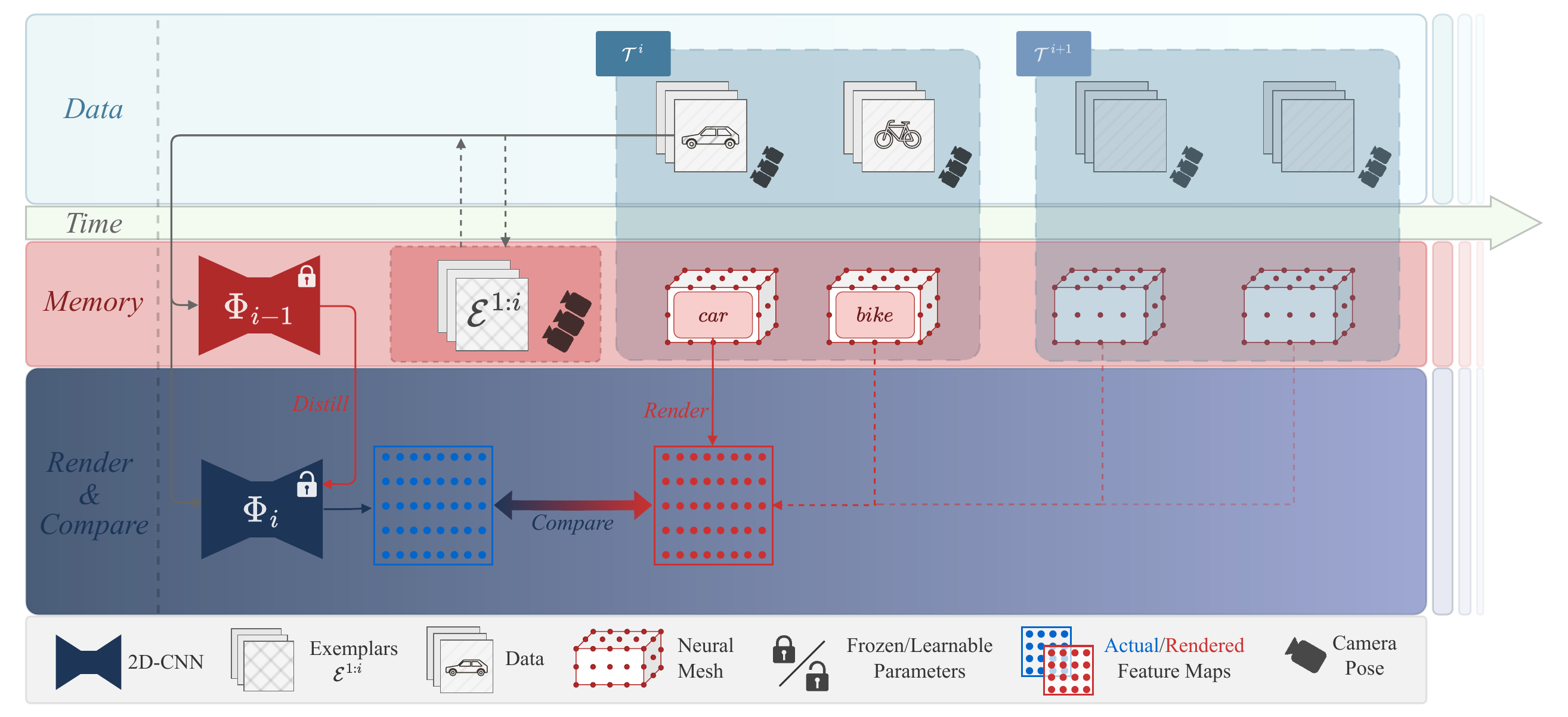}
    \caption{\textbf{We present iNeMo} that can perform class-incremental learning for pose estimation and classification, and performs well in out-of-distribution scenarios. 
    Our method receives tasks $\mathcal{T}^i$ over time that consist of images with camera poses for new classes. 
    We build up on Neural Mesh Models (NeMo)~\cite{wang_nemo} and abstract objects with simple cuboid 3D meshes, where each vertex carries a neural feature. The neural meshes are optimized together with a 2D feature extractor $\Phi_i$ and render-and-compare can then be used to perform pose estimation and classification. We introduce a memory that contains an old feature extractor $\Phi_{i-1}$ for distillation, a replay buffer $\mathcal{E}^{1:(i-1)}$ and a growing set of neural meshes $\mathfrak{N}$. Our results show that iNeMo outperforms all baselines for incremental learning and is significantly more robust than previous methods. 
    }
    \label{fig:teaser}
\end{figure}


Humans inherently learn in an incremental manner, acquiring new concepts over time, with little to no forgetting of previous ones. In contrast, trying to mimic the same behavior with machine learning suffers from \emph{catastrophic forgetting}~\cite{McCloskey1989Catastrophic,McRae1993Catastrophic,Kirkpatrick2017Overcoming}, where learning from a continual stream of data can destroy the knowledge that was previously acquired. In this context, the problem was formalized as \emph{class-incremental learning} and a variety of approaches have been proposed to address catastrophic forgetting for models that work in-distribution~\cite{LiH2018LwF, Douillard2020PODNet, Hou2019LUCIR, Rebuffi2017iCaRL, Liu2020Mnemonics,Liu2021RMM}. However, extending these methods to effectively manage out-of-distribution (OOD) scenarios~\cite{zhao2022ood} to the best of our knowledge has not been investigated.

Neural mesh models~\cite{wang_nemo} embed 3D object representations explicitly into neural network architectures,  and exhibit strong performance in generalizing to such OOD scenarios for classification and 3D pose estimation. However, as they consist of a 2D feature extractor paired with a generative model, their extension to a continual setting with existing techniques is not straight forward.
If one would only apply those techniques to the feature extractor, the previously learned neural meshes would become inconsistent and the performance of the model would drop. 

In this paper, we therefore present a strategy to learn neural mesh models incrementally  and refer to them as incremental Neural Mesh Models (iNeMo). As shown in Figure~\ref{fig:teaser}, 
in addition to the conventional techniques of knowledge distillation and maintaining a replay buffer, our approach introduces a memory that contains a continuously growing set of meshes that represent  object categories. To establish the learning of the meshes in an incremental setting, we extend the contrastive learning from~\cite{wang_nemo} by a latent space initialization strategy that enables us to allocate feature space for future unseen classes in advance, and a positional regularization term that forces the features of the different classes to consistently stay in respective latent space regions.
Through extensive evaluaitons on the Pascal3D~\cite{xiang2014beyond} and ObjectNet3D~\cite{xiang2016objectnet3d} datasets, we demonstrate that our method outperforms existing continual learning techniques and furthermore surpasses them by a large margin for out-of-distribution samples. Overall, our work motivates future research on joint 3D object-centric representations. 
In summary, the contributions of our work are: 
\begin{enumerate}
\item For the first time, we adapt the conventional continual learning techniques of knowledge distillation and replay to the 3D neural mesh setting. 
\item We propose a novel architecture, that can grow by adding new meshes for object categories over time.
\item To effectively train the features of the meshes, we introduce a strategy to partition the latent space and maintain it when new tasks are integrated. 
\item We demonstrate that incremental neural mesh models can outperform 2D baselines that use existing 2D continual learning techniques by $2-6\%$ in the in-domain and by $6-50\%$ in the OOD setting. 
\item Finally, we introduce the first incremental approach for pose estimation and show that the neural mesh models outperform 2D baselines. 
\end{enumerate}

\section{Related Work}
\subsection{Robust Image Classification and Pose Estimation}

\noindent \textbf{Image Classification}
has always been a cornerstone of computer vision. Groundbreaking models such as ResNets \cite{he2016deep}, Transformers~\cite{vaswani2017attention}, and Swin Transformers~\cite{liu2021swin} have been specifically designed for this task. However, these models predominantly target the in-distribution setting, leading to a significant gap in performance when faced with challenging benchmarks that involve synthetic corruptions~\cite{hendrycks2019benchmarking}, occlusions \cite{wang2020robust}, and out-of-distribution (OOD) images \cite{zhao2022ood}. Attempts to close this performance gap have included data augmentation \cite{hendrycks2019augmix} and innovative architectural designs, such as the analysis-by-synthesis approach \cite{kortylewski2020compositionalijcv}. 
Along this line of research, recently neural mesh models emerged as a family of models\cite{wang_nemo,wang2024neural,Ma20226DNeMo,wang2021neural} that learn a 3D pose-conditioned model of neural features and predict 3D pose and object class \cite{jesslen2023robust} by minimizing the reconstruction error between the actual and rendered feature maps using render-and-compare. 
Such models have shown to be significantly more robust to occlusions and OOD data. However, they can so far only be trained on fixed datasets.
In this work, we present the first approach to learn them in a class-incremental setting. 




\noindent \textbf{Object Pose Estimation} has been approached primarily as a regression problem \cite{Tulsiani2015Viewpoints, Mousavian20173D} or through keypoint detection and reprojection \cite{zhou2018starmap} in early methods. More recent research \cite{iwase2021repose, li2018deepim} addresses object pose estimation in complex scenarios like partial occlusion. NeMo~\cite{wang_nemo} introduces render-and-compare techniques for category-level object pose estimation, showcasing enhanced robustness in OOD conditions.
Later advancements in differentiable rendering~\cite{wang2022voge} and data augmentation~\cite{kouros2022category} for NeMo have led to further improvements in robust category-level object pose estimation, achieving state-of-the-art performance.
However, these approaches are confined to specific object categories and are designed for fixed training datasets only. In contrast, our method for the first time extends them to the class-incremental setting. 

\subsection{Class-Incremental Learning}
Class-incremental learning (also known as continual learning~\cite{De2021ContinualSurvey,Aljundi2019TaskFree,LopezPazR2017GEM} and lifelong learning~\cite{Aljundii2017ExpertGate,Chen2018Lifelong,Chaudhry2019AGEM}) aims at learning models from sequences of data. The foundational work of~\cite{Rebuffi2017iCaRL, castro2018end} replays exemplary data from previously seen classes. The simple strategy has inspired successive works~\cite{chaudhry2018riemannian, wu2018memory}. However, for such methods, sampling strategies and concept drift can impact overall performance. As a mitigation, more recent methods~\cite{Hou2019LUCIR,wu2019large} combine replay with other notable regularization schemes like knowledge distillation~\cite{LiH2018LwF}. In general, class-incremental methods leverage one or more principles from the following three categories:
(1) exemplar replay methods build a reservoir of samples from old training rounds~\cite{Rebuffi2017iCaRL,Shin2017GenerativeReplay,Liu2020Mnemonics,Prabhu2020GDumb,Bang2021Rainbow,Liu2024PlaceboCIL,Luo2023CVPR} and replay them in successive training phases as a way of recalling past knowledge,
(2) regularization-based (distillation-based) methods try to preserve the knowledge captured in a previous version of the model by matching logits~\cite{LiH2018LwF,Rebuffi2017iCaRL}, feature maps~\cite{Douillard2020PODNet}, or other information~\cite{Tao2020Topology,Wang2022FOSTER,Simon2021Learning,Joseph2022Energy,Pourkeshavarzi2022Looking,Liu2023Online} in the new model, and 
(3) network-architecture-based methods~\cite{Liu2020AANets,Wang2022FOSTER} design incremental architectures by expanding the network capacity for new class data or freezing partial network parameters to retain the knowledge about old classes.

In our work, we make use of principles from all three of the above by leveraging a replay memory, presenting a novel regularization scheme and adding newly trained neural meshes to the model over time. To the best of our knowledge, 
our method is the first to combine a 3D inductive bias with these strategies. 


\section{Prerequisites}

\subsection{Class Incremental Learning (CIL)}
Conventionally, classification models are trained on a single training dataset $\mathcal{T}$ that contains all classes.
Multi-class incremental learning departs from this setting by training models on sequentially incoming datasets of new classes that are referred to as tasks $\mathcal{T}^1, \mathcal{T}^2,...,\mathcal{T}^{N_{task}}$, where each task may contain more than one new class. After training on a new task $\mathcal{T}^i$, the model may be evaluated on a test dataset $\mathcal{D}^{1:i}$ that contains classes from all tasks up to $i$. 

When being trained on new tasks through a straightforward fine-tuning, models suffer from \emph{catastrophic forgetting}\cite{Kirkpatrick2017Overcoming}, which leads to bad performance on the previously seen classes. An intuitive approach to mitigate this effect is to use a \emph{replay buffer}\cite{Rebuffi2017iCaRL} that stores a few examplars $|\mathcal{E}^{i}|\ll |\mathcal{T}^i|$ from previous tasks and includes them with training data of the new task. Another common technique is \emph{knowledge distillation}~\cite{Hinton2015KD,LiH2018LwF} that keeps a copy of the model before training on the new task and ensures that distribution of the feature space from the old and new models are similar when presented the new data.

\subsection{Neural Mesh Models}
\label{nemo}
 
Neural mesh models combine a 2D feature extractor with generative 3D models, as shown by Wang et al.~\cite{wang_nemo} in their \href{https://openreview.net/pdf?id=pmj131uIL9H}{Figure 1}. 
The generative models are simple 3D abstractions in the form of cuboids for each class $c$ that are represented as meshes $\mathfrak{N}_c=(\mathcal{V}_c, \mathcal{A}_c, \Theta_c)$,  where $\mathcal{V}_c$ denotes the vertices, $\mathcal{A}_c$ denotes the triangles and $\Theta_c$ denotes the neural vertex features. The meshes are additionally accompanied by a set of background features $\mathcal{B}$. 
Given camera intrinsics and extrinsics, a mesh can then be rendered to a 2D feature map. 
The 2D feature extractor is usually a 2D CNN $\Phi(I)$ that takes the image as input to extract a feature map and is shared among all classes~\cite{jesslen2023robust}.  
Render-and-compare can then be used to check if the features rendered from the mesh align with the features extracted from the image to perform pose estimation~\cite{wang_nemo} or classification~\cite{jesslen2023robust}. 
We denote a normalized feature vector at vertex $k$ as $\theta_c^k$, its visibility in the image as $o_c^k$, its projected integer image coordinates as $\pi_c(k)$, and $f_{\pi_c(k)}$ as the normalized feature vector from the 2D feature extractor that corresponds to the rendered vertex $k$. 

During training, images and object poses are provided, and the vertex features $\Theta$, background features $\mathcal{B}$, and the 2D feature extractor $\Phi$ are trained. 
We model the probability distribution of a feature $f$ being generated from a vertex $v_c^k$ by defining $P(f|\theta_c^k)$ using a von Mises-Fisher (vMF) distribution to express the likelihood: 
\begin{equation}
\label{eq:vonmises}
P(f|\theta_c^k, \kappa)=C(\kappa)e^{\kappa (f^{\top}\cdot\theta_c^k)}\,\,,
\end{equation}
with mean $\theta_c^k$, concentration parameter $\kappa$, and normalization constant $C(\kappa)$~\cite{wang_nemo}.
In the next step, the extracted feature $f_{\pi_c(k)}$ is inserted into $P(f|\theta_c^k, \kappa)$ and maximized using contrastive learning. Simultaneously, the likelihood of all other vertices and background features is minimized: 
\begin{align}
    &\max \hspace{0.85cm} P(f_{\pi_c(k)}|\theta_c^k, \kappa), \label{eq:contrastiveproblem1}\\
    &\min \sum_{\theta^m\in\Bar{\theta}_c^k}P(f_{\pi_c(k)}|\theta^m, \kappa), \label{eq:contrastiveproblem2}
\end{align}
where the alternative vertices are defined as $\Bar{\theta}_c^k=\{\mathcal{B}\cup \Theta_{\Bar{c}}\cup (\Theta_c\setminus\mathcal{N}_c^k)\}$ with the neighborhood $\mathcal{N}_c^k=\{\theta_i \,|\, \lVert v_i^k - v^k_c\rVert < R \mathrm{\, \wedge \,} v_i^k\in \mathcal{V}_c \setminus v^c_k\}$ around $v_c^k$  determined by some pre-defined distance threshold $R$.
We formulate the Equations~\ref{eq:contrastiveproblem1} and~\ref{eq:contrastiveproblem2} into a single loss by taking the negative log-likelihood: 
\begin{equation}
\label{eq: nemoloss}
     \mathcal{L}_{\text{train}} = -\sum_{k}o^k_c \cdot \log(\frac{e^{\kappa (f_{\pi_c(k)}^\top \cdot \theta_c^k)}}{\sum_{\theta_m\in\Bar{\theta}_c^k}e^{\kappa (f_{\pi_c(k)}^\top \cdot \theta_m)}}),
\end{equation}
where considering $\kappa$ as a global hyperparameter allows cancelling out the normalization constants $C(\kappa)$.

The concentration parameter $\kappa$ determines the spread of the distribution and can be interpreted as an inverse temperature parameter.
In practice, the neural vertex features $\Theta$ and the background features $\mathcal{B}$ are unknown and need to be optimized jointly with the feature extractor $\Phi$.
This makes the training process initially ambiguous, where a good initialization of $\Phi$ and $\Theta$ is critical to avoid divergence. After each update of $\Phi$, we therefore follow Bai \textit{et al.}~\cite{bai2020coke} and use the momentum update strategy to train the foreground model $\Theta_c$ of a class $c$, as well as the background model $\mathcal{B}$:
\begin{equation}
\label{eq:momentum}
    \theta_c^{k,new} \xleftarrow{} o_c^k (1 - \eta) \cdot f_{\pi_c(k)} + (1 - o_c^k + \eta \cdot o_c^k) \theta_c^k\,\,,
\end{equation}
where $\eta$ is the momentum parameter.
The background model $\mathcal{B}$ is updated by sampling $N_{bgupdate}$ feature vectors at pixel positions that are not matched to any vertex of the mesh and replace the $N_{bgupdate}$ oldest features in $\mathcal{B}$.
Both $N_{bgupdate}$ and $\eta$ are hyperparameters.
For a more detailed description of this process, we refer to the supplementary material.

\begin{figure}[t!]
    \centering
    \includegraphics[width=1\textwidth]{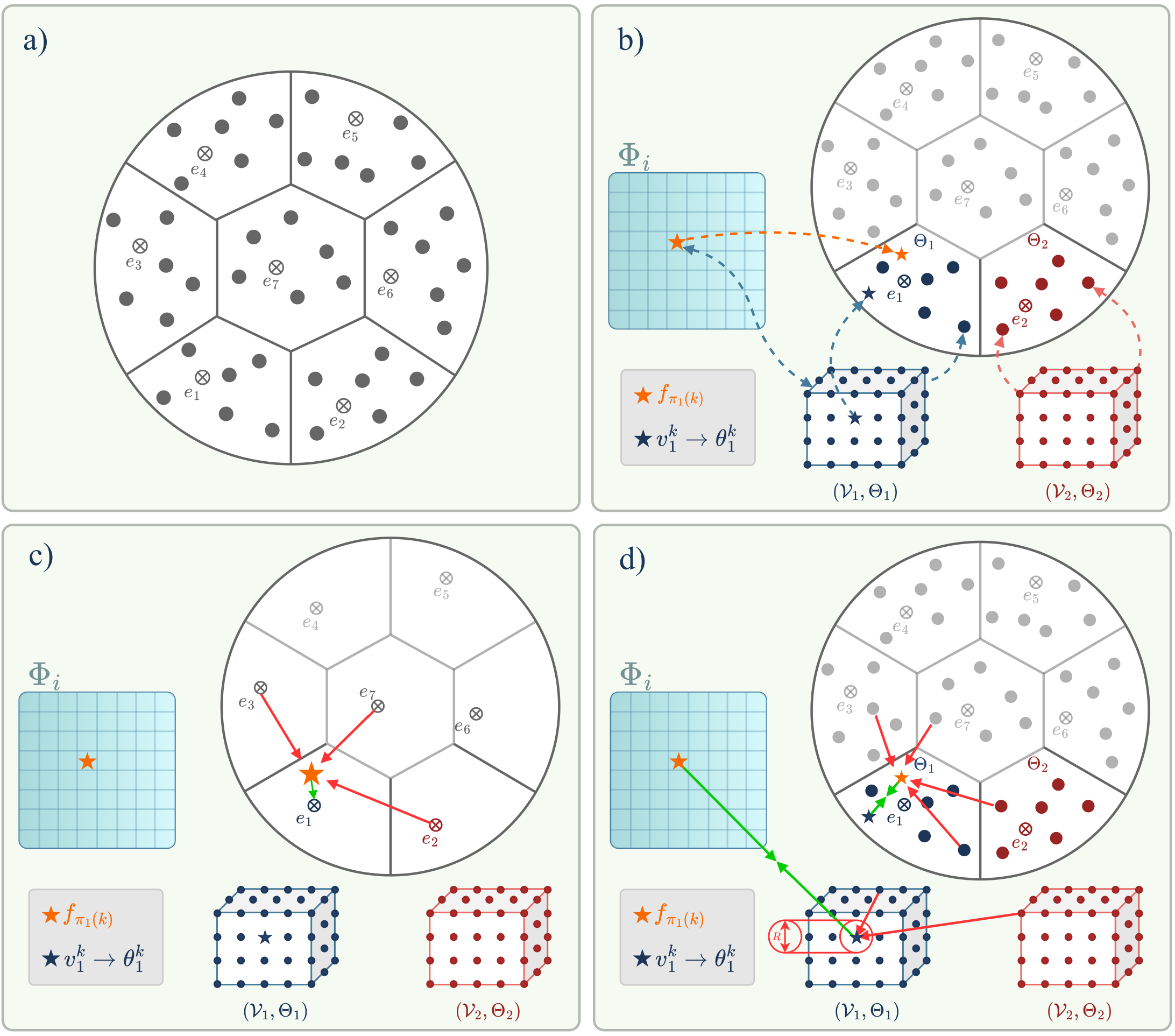}
    \caption{
    \textbf{Overview of Regularization:} 
    \textbf{a)} The features are constrained to lie on a unit sphere 
    and the latent space is initially uniformly populated. Centroids $e_i$ are then computed to lie maximally far apart, and the feature population is partitioned for a maximum number of classes. 
    \textbf{b)} When starting a new task, the vertex features for each new cube from this task are randomly initialized from some class partition.
    By projecting the locations of the vertices to images, corresponding image features are determined as illustrated by the orange star. 
    \textbf{c)} To avoid entanglement, we regularize the latent space by constraining the image feature to stay within the class partition using $\mathcal{L}_{etf}$.
    \textbf{d)} We then employ the contrastive loss $\mathcal{L}_{\text{cont}}$ that pulls the vertex and image features together and separates the image feature from other features of its own, and the other meshes. }
    \label{fig:latentspace}
\end{figure}

\section{Incremental Neural Mesh Models (iNeMo)}


Our goal is to learn a model that generalizes robustly in OOD scenarios, while being capable of performing class-incremental learning. To achieve this, we build up on neural mesh models~\cite{wang_nemo} and present a novel formulation for
class-incremental learning for classification and object pose estimation that we call iNeMo. An overview is provided in Figure~\ref{fig:teaser}. 


\subsubsection{Challenges in CIL.}
In the non-incremental setting, the contrastive loss in Equation \ref{eq: nemoloss} does not explicitly enforce separating classes, although in practice it is observed that the classes are separated well and accurate classification can be achieved~\cite{jesslen2023robust}. 
A naive extension of neural mesh models to class-incremental learning is to simply add a mesh $\mathfrak{N}_c$ for each new class. However, the challenge lies in updating the shared 2D feature extractor $\Phi$. If adding classes naively, achieving a discriminative latent space requires restructuring it as a whole and therefore implies significant changes in both, the CNN backbone $\Phi$ and the neural meshes $\Theta$, leading to catastrophic forgetting if no old training samples are available or other measures are taken. 
Therefore, in the following we present a novel class-incremental learning strategy that maintains a well structured latent space from the beginning.


\subsection{Initialization}

\subsubsection{Latent Space.}

As the features $\theta^k_c$ are normalized, they lie on a unit sphere. We therefore define an initial population $\bm{H}=\{\bm{h}_j \,|\, \bm{h}_j\in \R^d \,\wedge\, \lVert \bm{h}_j\rVert=1 \}$ of the latent space for all vertices and classes by uniformly sampling the sphere.
To partition the latent space, we define a fixed upper bound of classes $N$. We then generate centroids $\bm{E}=[\bm{e}_1,...,\bm{e}_{N}]$ for all the classes on the unit sphere that are pairwise maximally far apart by solving the equation for a simplex Equiangular Tight Frame (ETF)~\cite{papyan2020prevalence}: 
\begin{equation}
\label{eq: ETF}
    \bm{E}=\sqrt{\frac{N}{N-1}}\bm{U}(\bm{I}_N-\frac{1}{N}\bm{1}_N\bm{1}_N^\top) \,\,, 
\end{equation}
where $\bm{I}_N$ denotes the $n-$dimensional identity matrix, $\bm{1}_n$ is an all-ones vector, and $\bm{U}\in\R^{d\times n}$ is any matrix that allows rotation. The column vectors are of equal Euclidean norm and any pair has an inner product of $\bm{e}_i^{\top} \cdot \bm{e}_j=-\frac{1}{N-1}$ for $i\neq j$, which together ensures pairwise maximum distances. Finally, we assign the features $\bm{h}_j$ to classes by determining the respective closest centroid from $\bm{E}$, which leads to a partitioning ${\bm{H}_1,...,\bm{H}_{N}}$ of $\bm{H}$. An illustration of this strategy is provided in Figure~\ref{fig:latentspace} a).

\subsubsection{Task $\mathcal{T}^i$.} 
At the start of each task, we need to introduce new neural meshes. 
Following Wang \textit{et al.}~\cite{wang_nemo}, for each new class $c$ we initialize $\mathfrak{N}_c$ as a cuboid where its dimensions are determined from ground-truth meshes and vertices are sampled on a regular grid on the surface. 
As illustrated in Figure~\ref{fig:latentspace} b), we then pick the partition $\bm{H}_{c}$ of initial features and randomly assign them to the vertices of the new mesh  $\Theta_c$. 
We initialize the feature extractor $\Phi_0$ with unsupervised pre-training using DINO-v1~\cite{caron2021emerging}.  
As shown in Figure~\ref{fig:teaser}, to train for a new task, we make a copy $\Phi_{i} = \Phi_{i-1}$ and then leverage $\Phi_{i-1}$ for knowledge distillation.
If available, we discard any old network $\Phi_{i-2}$.

\subsection{Optimization}

\subsubsection{Positional Regularization.}
To ensure that our latent space maintains the initial partitioning over time, 
we introduce a penalty of the distance of the neural features $\Theta_c$ to their  corresponding class centroid $\bm{e}_c$:
\begin{equation}
\label{eq: etfReg}
\mathcal{L}_{\text{etf}} = -\sum_{k}o_c^k \cdot \log\left(\frac{e^{\kappa_2(f_{\pi_c(k)}^\top \cdot \bm{e}_c)}}
{
     \sum_{\bm{e}_m \in \bm{E}}e^{\kappa_2(f_{\pi_c(k)}^\top \cdot \bm{e}_m)}}\right).
\end{equation}
This is illustrated in Figure~\ref{fig:latentspace} c). 

\subsubsection{Continual Training Loss.}
We denote any unused partitions in $\bm{H}$ with $\Bar{\bm{H}}$ and limit the spread of the neural meshes in the current task to refrain from $\Bar{\bm{H}}$ by posing the following additional contrastive loss: 
\begin{equation}
\label{eq: continualLoss}
     \mathcal{L}_{\text{cont}} = -\sum_{k}o_c^k \cdot \log\Biggl(\frac{e^{\kappa_1 (f_{\pi_c(k)}^\top \cdot \theta_c^k)}}{
     \sum_{\theta^m\in\Bar{\theta}_k}e^{\kappa_1 (f_{\pi_c(k)}^\top \cdot \theta^m)}
     + 
     \sum_{h_j\in \Bar{\bm{H}}}e^{\kappa_1 (f_{\pi_c(k)}^\top \cdot h_j)}
     }\Biggr).
\end{equation}
The denominator is split into two parts, where the first one minimizes Equation~\ref{eq:contrastiveproblem2} and the second part corresponds to the additional constraint imposed by the features in the unused partitions $\Bar{\bm{H}}$. This is illustrated in Figure~\ref{fig:latentspace} d).

\subsubsection{Knowledge Distillation.}
To mitigate forgetting, as indicated in Figure~\ref{fig:teaser}, we additionally use a distillation loss after the initial task.
The new inputs are also fed through the frozen backbone $\Phi_{i-1}$ of the previous task to obtain its feature map. Specifically, let $\hat{f}_{\pi_c(k)}$ denote the old feature for the  vertex $k$. 
To distill classes from previous tasks into $\Phi_{i}$, we formulate the distillation using the Kullback-Leibler divergence:
\begin{equation}
\label{eq: kdReg}
\mathcal{L}_{\text{kd}} = -\sum_k\sum_m p_m(\hat{f}_{\pi_c(k)})\log(\frac{p_m(\hat{f}_{\pi_c(k)})}{p_m(f_{\pi_c(k)})}),
\end{equation}
where:
\begin{equation}
\label{eq: pReg}
     p_m(f_{\pi_c(k)}) = \frac{e^{\kappa_{3}(f_{\pi_c(k)}^\top \cdot \theta_m)}}{
     \sum_{\theta_m \in \Theta^{i-1}} e^{\kappa_{3}(f_{\pi_c(k)}^\top \cdot \theta_m)}
     }.
\end{equation}
Note that, unless we are considering an exemplar of a previous task, the real corresponding feature $\theta_c^k$ is not even considered in this formulation.
However, the aim here is not to optimize $\Phi_{i}$ for the current task, but to extract the dark knowledge~\cite{Hinton2015KD} from $\Phi_{i-1}$ about classes from previous tasks.
Consequently, the concentration $\kappa_3<1$ has to be small to get usable gradients from all likelihoods.

\subsubsection{Continual Training.}
During training of $\Phi_i$, we optimize the combined training objective:
\begin{equation}
    \label{eq: trainLoss}
    \mathcal{L} = \mathcal{L}_{\text{cont}} + \lambda_{\text{etf}}\mathcal{L}_{\text{etf}} + \lambda_{\text{kd}}\mathcal{L}_{\text{kd}},
\end{equation}
where $\lambda_{\text{etf}}$ and $\lambda_{\text{kd}}$ are weighting parameters.

\subsection{Exemplar Selection}

At each training stage, we randomly remove exemplars from old classes to equally divide our replay buffer for the current number of classes. Xiang \textit{et al.}\cite{xiang2014beyond} showed that certain classes are heavily biased towards certain viewing angles. 
Therefore, to increase the robustness and accuracy for rarely appearing view directions,
we propose an exemplar selection strategy that takes viewing angles into account. 
Assuming we want to integrate a new class and the available slots for it are $m$, we build a $b$-bin histogram across the azimuth angles 
and randomly select $\lfloor m/b \rfloor$ exemplars for each bin. When insufficient exemplars are available for a bin we merge it together with a neighboring one. In case the process yields less than $m$ exemplars in total, we fill up remaining slots with random samples. 
When reducing the exemplar sets, we evenly remove samples from each bin to maintain the balance across the azimuth angle distribution.



\subsection{Inference}

\subsubsection{Classification.}
\label{Inference_Classification}
Following Jesslen \textit{et al.}~\cite{jesslen2023robust}, we perform classification via a vertex matching approach.
For each feature $f_i$ in the produced feature map of $\Phi$, we compute its similarities to the foreground ($\Theta$) and background ($\mathcal{B}$) models.
We define the background score $s^i_\beta$ and the class scores $s^i$ for each class $c$ as
\begin{align}
    s^i_c&=\max_{\theta^l_c\in \Theta_c} f_i^\top \cdot \theta^l_c,\\
    s^i_\beta&=\max_{\beta^l\in \mathcal{B}} f_i^\top \cdot \beta^l,
\end{align}
where we identify a feature as being in the foreground $\mathcal{F}$, if there is at least one $s^i_c>s^i_\beta$ and classify based on the foreground pixels only.

In contrast to Jesslen \textit{et al.}~\cite{jesslen2023robust}, we additionally include an uncertainty term to reduce the influence of features that can not be identified with high confidence.
In the following, we denote the $n-$th largest class score for feature $f_i$ as $max^{(n)}_{S^i}$. 
The final score of class $y$ is then given as 
\begin{equation}
    s_y = \max_{i\in\mathcal{F}} \left[s_y^i-(1-(max^{(1)}_{s^i}-max^{(2)}_{s^i}))\right],
\end{equation}
where the subtracted term indicates a measure of confusion estimated based on the difference of the two highest class scores for foreground feature $f_i$.
The predicted category is then simply the class $c$ that maximizes this score.

\subsubsection{Pose Estimation.}
For pose estimation we use the same render-and-compare approach as Wang \textit{et al.}~\cite{wang_nemo} together with the template matching proposed by Jesslen \textit{et al.}~\cite{jesslen2023robust} for speedup.
For more information about the pose estimation, we refer the reader to the supplemental material.

\section{Experiments}


In the following, we explain the experimental setup and then discuss the results of our incremental neural mesh models for image classification and 3D pose estimation on both, in-domain and OOD datasets. 
For a comprehensive ablation study of all components of our model, we refer to the supplemental material.


\subsection{Datasets and Implementation Details}


\noindent \subsubsection{In-Domain-Datasets.} 
\textbf{PASCAL3D+}~\cite{xiang_wacv14} ({P3D}) has high-quality camera pose annotations with mostly unoccluded objects, making it ideal for our setting.
However, with only $12$ classes it is small compared to other datasets used in continual learning~\cite{krizhevsky2009learning, russakovsky2015imagenet}.
\textbf{ObjectNet3D}~\cite{xiang2016objectnet3d} (O3D) contains $100$ classes and presents a significantly more difficult setting.
Camera pose annotations are less reliable and the displayed objects can be heavily occluded or truncated, making both the vertex mapping and the update process noisy.

\subsubsection{OOD-Datasets.}

The \textbf{Occluded-PASCAL3D+}~\cite{wang2020robust}  ({O-P3D}) and \textbf{corrupted-PASCAL3D+} ({C-P3D}) datasets are variations of original P3D and consist of a test dataset only.
In the O-P3D dataset, parts of the original test datasets have been artificially occluded by superimposing occluders on the images with three different levels: L1 ($20\%-40\%$), L2 ($40\%-60\%$) and L3 ($60\%-80\%$).
The C-P3D dataset, on the other hand, follows~\cite{hendrycks2019benchmarking} and tests robustness against image corruptions.
We evaluate 19 different corruptions with a severity of $4$ out of $5$, using the \verb|imagecorruptions|~\cite{imagecorruptions} library. Finally, we consider the \textbf{OOD-CV}~\cite{zhao2022ood} dataset, which provides a multitude of severe domain shifts.

\noindent \subsubsection{Implementational Details.}
We choose a ResNet50 architecture for our feature extractor $\Phi$ with two upsampling layers and skip connections, resulting in a final feature map at $\frac{1}{8}$ of the input resolution.
Each neural mesh $\mathfrak{N}_y$ contains approximately $1,100$ uniformly distributed vertices with a neural texture of dimension $d=128$.
We train for $50$ epochs per task, with a learning rate of $1e-5$ that is halved after $10$ epochs.
The replay buffer can store up to $240$ and $2,000$ samples for P3D and O3D respectively.
Our feature extractor is optimized using Adam with default parameters and the neural textures $\Theta$ are updated with momentum of $\eta=0.9$.
During pose estimation, we initialize the camera pose using template matching as proposed by~\cite{jesslen2023robust} and optimize it with PyTorch3D's differentiable rasterizer~\cite{ravi2020pytorch3d}.
The initial camera pose is refined by minimizing the reconstruction loss between the feature map produced by $\Phi$ and the rendered mesh.
We use Adam with a learning rate of $0.05$ for $30$ total epochs and a distance threshold $R=48$ to measure the neighborhood $\mathcal{N}_c^k$ in Equation~\ref{eq: continualLoss}. 
Each term in the combined loss in Equation~\ref{eq: trainLoss} is assigned a weighting and concentration parameter.
The weighting parameters are $\lambda_{\text{etf}}=0.2$ and $\lambda_{\text{kd}}=2.0$ and as concentration parameters we choose $\kappa_1=1/0.07\approx 14.3$, $\kappa_2=1$, and $\kappa_3=0.5$.
We provide further details on the training settings of the baselines in the supplemental material. 


{
\setlength{\tabcolsep}{1.8mm}{
\begin{table*}[tp]
  \small
  \centering
    \caption{
  Average classification accuracies on Pascal3D (P3D) and ObjectNet3D (O3D). Training data has been split into a base task (denoted $Bn$ for size $n$) and evenly sized increments (denoted $+n$ for size $n$). As visible, our method consistently outperforms the baselines by a significant margin.
  }
  \begin{tabu}{llcccccccc}
  \toprule
  \multirow{2.5}{*}{Metric} &  \multirow{2.5}{*}{Method} & \multirow{2.5}{*}{Repr.}& & \multicolumn{2}{c}{\emph{P3D}} & & &\emph{O3D}&\\
  \cmidrule{5-6} \cmidrule{8-10}
  \rowfont{\tiny}&     &  & & $B0+6$ & $B0+3$  &   & $B0+20$& $B0+10$&$B50+10$ \\
    \midrule
    &  LWF & R50.  && 93.83 & 89.34 & & 67.78 &48.82 &46.73 \\
    &  FeTrIL & R50.  && 95.64 & 96.82 & & 67.18 &70.34 &70.43 \\
    Classification&  FeCAM & R50.  && 84.85 & 64.36 & & 67.96 &69.59 &72.21 \\
    \multirow{1}{*}{$acc(1\,{\colon}i)$ in \% $\uparrow$
    }&  iCaRL & R50 && 97.1 &  93.80 && 72.55 & 57.46&64.02\\
    &  DER& R50 && 96.69 &  94.18 && 78.55 & 76.33&75.17\\
    &  Podnet& R50 && 95.13  & 91.71 & & 71.96 & 65.21&72.98\\
    &  Ours & NeMo && \bf{98.82} & \bf{98.21} & & \bf{89.25} & \bf{88.85}&\textbf{84.20}\\
  \bottomrule
    \end{tabu}
  \label{table: classP3DO3D}
\end{table*}
}
}

\noindent \subsubsection{Evaluation.}

We evaluate our method and its baselines on both, class-incremental classification and class-incremental pose estimation.
For the methods trained on P3D, we evaluate on the P3D test dataset, the O-P3D dataset, and the C-P3D dataset.
When training on O3D or OOD-CV, we evaluate on their corresponding test dataset only. 
For classification, we follow previous work~\cite{Rebuffi2017iCaRL,LiH2018LwF,Liu2020Mnemonics} and consider the mean accuracy over all tasks $acc(1\,{\colon}i)$
of $\Phi_i$, after training on $\mathcal{T}^i$ on test dataset $\mathcal{D}$ for classes $1..i$.
The 3D pose of an object can be represented with azimuth, elevation, and roll angle.
We measure the deviation of predicted and ground-truth pose in terms of these angles according to the error of the predicted and the ground-truth rotation matrix $\Delta(R_{\text{pred}},R_{\text{gt}})=\lVert \log m(R^\top_{\text{pred}}R_{\text{gt}})\rVert_F / \sqrt{2}$ as proposed by~\cite{zhou2018starmap}.
Following previous work~\cite{zhou2018starmap,wang_nemo}, we report the accuracy according to the thresholds $\pi/6$ and $\pi/18$.

{\setlength{\tabcolsep}{0.7mm}{

\begin{table*}[tp]
  \small
  \centering
    \caption{
  Average classification and pose estimation accuracies on Pascal3D (P3D) and its variants.
  As visible, iNeMo outperforms all 2D baselines consistently for classification and by an especially large margin for the OOD and strong occlusion cases. We also present the first approach for incremental pose estimation and outperform other methods in most cases for $\pi/6$, while we consistently outperform them for the tighter error bound $\pi/18$. 
  Note that for all evaluations except OOD-CV, we use the model trained on 4 tasks that is also displayed in Figure~\ref{fig: clsLosses}.
  As OOD-CV provides a separate training set of 10 classes, we consider 2 tasks with 5 classes.
  }
  \begin{tabular}{llcccccccccccc}
  \toprule
  \multirow{2.5}{*}{Metric} & & \multirow{2.5}{*}{Method} & \multirow{2.5}{*}{Repr.} & \emph{P3D} & & \multicolumn{3}{c}{\emph{Occluded P3D}} & & \emph{C-P3D} && \emph{OOD-CV}\\
  \cmidrule{7-9} 
  &&&&   & & $L1$ & $L2$  & $L3$ & & & &  \\
   \midrule
   & & LWF & R50 & 89.34 & & 30.58 & 21.64 & 14.65 && 66.17 && 57.61\\
      & & FeTrIL & R50 & 96.82 & & 88.34 & 76.28 & 55.89 && 39.02 && 63.74\\
   \multirow{1}{*}{$acc(1\,{\colon}i)$ in \% $\uparrow$}& & FeCAM & R50 & 84.85 & & 52.53 & 42.75 & 34.28 && 42.38 && 56.05\\

   Classification& & iCaRL & R50 & 93.80 & & 34.95 & 26.00 & 16.93 & & 76.24 && 61.80\\
   
    & & DER & R50 & 94.18 & & 49.70 & 36.86 & 22.76 & & 69.56 && 56.35\\
       
    & & PODNet &R50& 91.91 & & 42.40 & 32.99 & 22.43 && 68.46 && 57.10\\
    & & Ours & NeMo& \bf{98.21} & & \bf{94.19} & \bf{87.20} & \bf{71.55} && \bf{83.09} && \bf{80.82}\\

  \midrule
   & & LwF & R50  & 53.47 & & 44.58 & 39.77 & \bf{36.61}  && 53.55 && 30.65 \\
   Pose $\pi / 6$
   & & iCaRL  & R50& 57.74 & & 44.03 & 38.15 & 33.52  & & \bf{54.57} && 28.71 \\
       \multirow{1}{*}{$acc(1\,{\colon}i)$ in \% $\uparrow$} 
   &  & Ours & NeMo & \bf{79.28} & & \bf{64.71} & \bf{52.26} & 34.01 && 47.30 && \bf{33.75}\\

     \midrule
    & & LwF  & R50 & 20.33 & & 12.03 & 8.38 &  5.52 && 17.29 && 8.04 \\
    Pose $\pi / 18$
   & & iCaRL  & R50 & 22.76 & & 11.04 & 7.33 & 4.56  && 17.81 && 8.04  \\
    \multirow{1}{*}{$acc(1\,{\colon}i)$ in \% $\uparrow$}    
   &  & Ours & NeMo &\bf{51.73} & & \bf{35.53} & \bf{26.88} & \bf{10.672} && \bf{23.02} &&\bf{12.8}\\
  \bottomrule
    \end{tabular}
  \label{table: robustness}
\end{table*}
}
}

\noindent \subsubsection{Baselines.}

For the task of class-incremental learning, we compare against a collection of replay-based and replay-free methods.
For the replay-based methods, we choose the seminal work iCaRL~\cite{Rebuffi2017iCaRL}, and the more recent PODNet~\cite{Douillard2020PODNet} and DER~\cite{yan2021dynamically}.
For replay-free methods, we choose the seminal work LwF~\cite{LiH2018LwF} and the two state-of-the-art methods FeTrIL~\cite{fetril} and FeCAM~\cite{fecam}.
All approaches are implemented using the PyCIL library~\cite{Zhou2023PyCIL} and trained with the original hyperparameters as in~\cite{Zhou2023PyCIL}. 
For a fair comparison of all methods, we use the ResNet-50 backbone initialized with DINO-v1\cite{caron2021emerging} pre-trained weights.

To the best of our knowledge, incremental pose estimation with a class-agnostic backbone has not been explored before.
We define incremental pose estimation baselines by discretizing the polar camera coordinates and formulate pose estimation as a classification problem following~\cite{zhou2018starmap}.
More specifically, we define $42$ bins for each azimuth, elevation and roll angle, making it a $42 \cdot 3=126$ class classification problem~\cite{zhou2018starmap}. 
This allows a straightforward extension of conventional class-incremental learning techniques to the setting of pose estimation.
We provide such class-incremental pose estimation results using the training procedure of iCaRL~\cite{Rebuffi2017iCaRL} and LwF~\cite{LiH2018LwF}.
Both methods are trained for $100$ epochs per task using SGD with a learning rate of $0.01$ as proposed by~\cite{zhou2018starmap}.



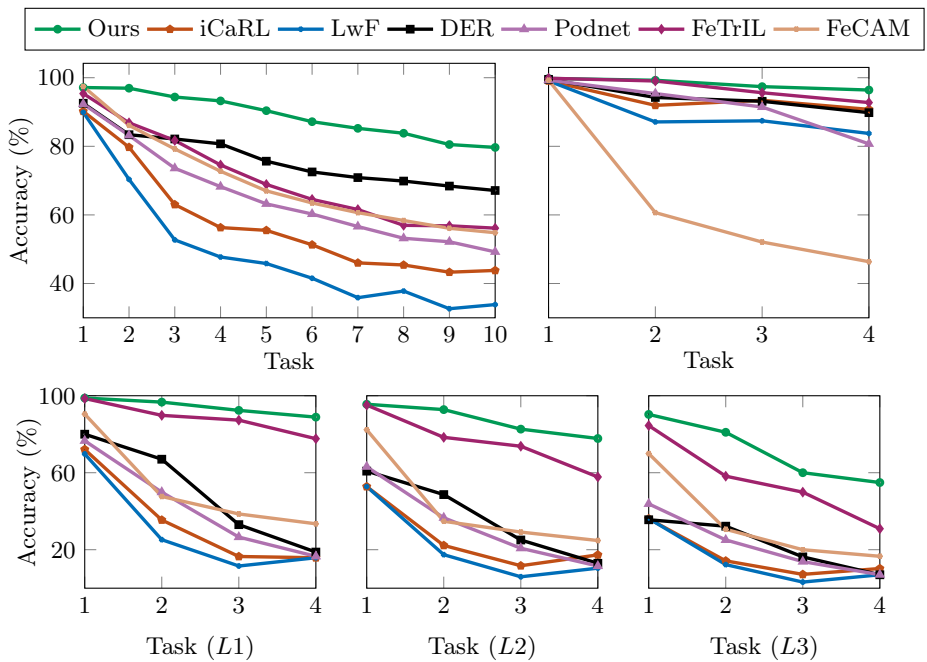
\begin{figure}[t]
\centering
\begin{subfigure}[b]{\textwidth}
\begin{center}
\resizebox{12.5cm}{!}{
    \begin{tikzpicture}
    \pgfplotstableread{data/cls_acc_o3d_CI10_new.tex}{\table}
    \pgfplotstableread{data/cls_acc_p3d_CI3_new.tex}{\table}
    \pgfplotsset{
        every axis plot/.append style={
            mark size=1pt
        }
    }
    \begin{axis}[
            at={(0,0)},
            axis lines=box,
            axis line style={->},
            xlabel={Task},
            ylabel={Accuracy (\%)},
            legend style={at={(0.95,1.05)},anchor=south, legend columns=7},
            x label style={at={(axis description cs:0.5,-0.1)},anchor=north},
            y label style={at={(axis description cs:-0.1,0.5)},anchor=south},
            xtick={0, 1, 2, 3, 4, 5, 6, 7, 8, 9, 10},
            xmax=10,
            xmin=1,
            ymin=30,
            x=0.6cm, 
            y=0.45cm/10,
    ]
            \addplot[ForestGreen, mark=o, line width=1.25pt] table [x={x}, y={Ours}] {\table};
            \addlegendentry{Ours}
            \addplot[Bittersweet, mark=pentagon,line width=1.25pt] table [x={x}, y={iCaRL}] {\table};
            \addlegendentry{iCaRL}
            \addplot[NavyBlue, mark=x,line width=1.25pt] table [x={x}, y={LwF}] {\table};
            \addlegendentry{LwF}   
            \addplot[black, mark=square,line width=1.25pt] table [x={x}, y={DER}] {\table};
            \addlegendentry{DER}
            \addplot[Orchid, mark=triangle,line width=1.25pt] table [x={x}, y={Podnet}] {\table};
            \addlegendentry{Podnet}     
            \addplot[RedViolet, mark=diamond,line width=1.25pt] table [x={x}, y={FeTrIL}] {\table};
            \addlegendentry{FeTrIL} 
            \addplot[Tan, mark=asterisk,line width=1.25pt] table [x={x}, y={FeCAM}] {\table};
            \addlegendentry{FeCAM} 
        \end{axis}
        \hfill
        \begin{axis}[
            at={(0.5\linewidth,0)},
            axis lines=box,
            axis line style={->},
            xlabel={Task},
            x label style={at={(axis description cs:0.5,-0.1)},anchor=north},
            xtick={1,2,3,4},
            yticklabels={},
            xmax=4,
            xmin=1,
            ymin=30,
            ymax=103,
            x=1.4cm, 
            y=0.45cm/10
    ]
            \addplot[ForestGreen, mark=o,line width=1.25pt] table [x={x2}, y={Ours2}] {\table};
            \addplot[Bittersweet, mark=pentagon,line width=1.25pt] table [x={x2}, y={iCaRL2}] {\table};
            \addplot[NavyBlue, mark=x,line width=1.25pt] table [x={x2}, y={LwF2}] {\table};
            \addplot[black, mark=square,line width=1.25pt] table [x={x2}, y={DER2}] {\table};
            \addplot[Orchid, mark=triangle,line width=1.25pt] table [x={x2}, y={Podnet2}] {\table};
            \addplot[RedViolet, mark=diamond,line width=1.25pt] table [x={x}, y={FeTrIL2}] {\table};
            \addplot[Tan, mark=asterisk,line width=1.25pt] table [x={x}, y={FeCAM2}] {\table};

        \end{axis}
    \end{tikzpicture}
}
\end{center}
\end{subfigure}
\begin{subfigure}[b]{\textwidth}
\begin{center}
\resizebox{12cm}{!}{
    \begin{tikzpicture}
    \pgfplotstableread{data/cls_acc_Occ_1.tex}{\table}
    \pgfplotstableread{data/cls_acc_Occ_2.tex}{\table}
    \pgfplotstableread{data/cls_acc_Occ_3.tex}{\table}
    \pgfplotsset{
        every axis plot/.append style={
            mark size=1pt
        }
    }
    
    \begin{axis}[
            at={(0,0)},
            axis lines=box,
            axis line style={->},
            xlabel={Task ($L1$)},
            ylabel={Accuracy (\%)},
            legend style={at={(0.75,1.05)},anchor=south, legend columns=5},
            x label style={at={(axis description cs:0.5,-0.2)},anchor=north},
            y label style={at={(axis description cs:-0.15,0.5)},anchor=south},
            xtick={1,2,3,4},
            ytick={20,60,100},
            xmax=4,
            xmin=1,
            ymin=0,
            ymax=100,
            x=1cm, 
            y=0.25cm/10
    ]
    
            \addplot[ForestGreen, mark=o, line width=1.25pt] table [x={xocc1}, y={Oursocc1}] {\table};
            \addplot[Bittersweet, mark=pentagon,line width=1.25pt] table [x={xocc1}, y={iCaRLocc1}] {\table};
            \addplot[NavyBlue, mark=x,line width=1.25pt] table [x={xocc1}, y={LwFocc1}] {\table};
            \addplot[black, mark=square,line width=1.25pt] table [x={xocc1}, y={DERocc1}] {\table};
            \addplot[Orchid, mark=triangle,line width=1.25pt] table [x={xocc1}, y={Podnetocc1}] {\table};
            \addplot[RedViolet, mark=diamond,line width=1.25pt] table [x={xocc1}, y={FeTrILocc1}] {\table};
            \addplot[Tan, mark=asterisk,line width=1.25pt] table [x={xocc1}, y={FeCAMocc1}] {\table};
        \end{axis}
        \hfill
    \begin{axis}[
            at={(0.3\linewidth,0)},
            axis lines=box,
            axis line style={->},
            xlabel={Task ($L2$)},
            x label style={at={(axis description cs:0.5,-0.2)},anchor=north},
            xtick={1,2,3,4},
            ytick={20,60,100},
            yticklabels={},
            xmax=4,
            xmin=1,
            ymin=0,
            ymax=100,
            x=1cm, 
            y=0.25cm/10
    ]
            \addplot[ForestGreen, mark=o, line width=1.25pt] table [x={xocc2}, y={Oursocc2}] {\table};
            \addplot[Bittersweet, mark=pentagon*,line width=1.25pt] table [x={xocc2}, y={iCaRLocc2}] {\table};
            \addplot[NavyBlue, mark=x,line width=1.25pt] table [x={xocc2}, y={LwFocc2}] {\table};
            \addplot[black, mark=square*,line width=1.25pt] table [x={xocc2}, y={DERocc2}] {\table};
            \addplot[Orchid, mark=triangle,line width=1.25pt] table [x={xocc2}, y={Podnetocc2}] {\table};
            \addplot[RedViolet, mark=diamond,line width=1.25pt] table [x={xocc2}, y={FeTrILocc2}] {\table};
            \addplot[Tan, mark=asterisk,line width=1.25pt] table [x={xocc2}, y={FeCAMocc2}] {\table};
        \end{axis}
         \hfill
    \begin{axis}[
            at={(0.6\linewidth,0)},
            axis lines=box,
            axis line style={->},
            xlabel={Task ($L3$)},
            x label style={at={(axis description cs:0.5,-0.2)},anchor=north},
            xtick={1,2,3,4},
            ytick={20,60,100},
            yticklabels={},
            xmax=4,
            xmin=1,
            ymin=0,
            ymax=100,
            x=1cm, 
            y=0.25cm/10
    ]
            \addplot[ForestGreen, mark=o, line width=1.25pt] table [x={xocc3}, y={Oursocc3}] {\table};
            \addplot[Bittersweet, mark=pentagon,line width=1.25pt] table [x={xocc3}, y={iCaRLocc3}] {\table};
            \addplot[NavyBlue, mark=x,line width=1.25pt] table [x={xocc3}, y={LwFocc3}] {\table};
            \addplot[black, mark=square,line width=1.25pt] table [x={xocc3}, y={DERocc3}] {\table};
            \addplot[Orchid, mark=triangle,line width=1.25pt] table [x={xocc3}, y={Podnetocc3}] {\table};
            \addplot[RedViolet, mark=diamond,line width=1.25pt] table [x={xocc3}, y={FeTrILocc3}] {\table};
            \addplot[Tan, mark=asterisk,line width=1.25pt] table [x={xocc3}, y={FeCAMocc3}] {\table};

        \end{axis}
    \end{tikzpicture}
}
\end{center}
\label{fig: occluded}
\end{subfigure}
\caption{
Comparison of classification performance decay over tasks for our method and the baselines. 
\textbf{Top-Left:} Results for O3D (100 classes) split into 10 even tasks. \textbf{Top-Right:} Results for P3D (12 classes) split into 4 even tasks. 
\textbf{Bottom:} Results for O-P3D with occlusion levels L1, L2 and L3 after each task. 
One can observe that our method outperforms all other methods. Especially in the occluded cases, our method outperforms them by a very large margin up to $70\%$, even still showing strong performance for the largest occlusion level L3 with $60 - 80\%$ occlusions. 
}
\label{fig: clsLosses}
\end{figure}


%





\subsection{Robust Class-Incremental Classification}
In Table~\ref{table: classP3DO3D}, we provide the in-distribution classification results for P3D and O3D. 
 Our method outperforms the baselines in all cases, including the harder O3D setting with $100$ classes.  Furthermore, Table~\ref{table: robustness} shows the comparison of class-incremental results on all P3D variants for both, classification and 3D pose estimation. As visible, our method outperforms the other methods with a large margin under domain shifts: for the L3 occluded case, it is larger than $48\%$, for the corrupted C-P3D it is larger than $6\%$, and for the OOD-CV dataset it is larger than $19\%$. Figure~\ref{fig: clsLosses} shows task-wise accuracy on the O3D/P3D dataset for 10 and 4 even tasks respectively, as well as the task-wise accuracy on the O-P3D dataset for all occlusion levels. The same observation as before can be made, where our method exhibits significantly less performance decay over new tasks. This overall demonstrates that our incremental neural mesh models outperform their baselines decisively in robustness.


\begin{figure}[t]
\centering
\begin{subfigure}[b]{\textwidth}
\begin{center}
\resizebox{11.5cm}{!}{
    \begin{tikzpicture}
    \pgfplotstableread{data/pe_acc_p3d_CI3.tex}{\table}
    \pgfplotstableread{data/pe_acc_p3d_pi18.tex}{\table}

    \pgfplotsset{
        every axis plot/.append style={
            mark size=1pt
        }
    }
    \begin{axis}[
            at={(0,0)},
            axis lines=box,
            axis line style={->},
            xlabel={Task},
            ylabel={Accuracy (\%)},
            legend style={at={(0.95,1.05)},anchor=south, legend columns=5},
            x label style={at={(axis description cs:0.5,-0.1)},anchor=north},
            y label style={at={(axis description cs:-0.1,0.5)},anchor=south},
            xtick={0, 1, 2, 3, 4},
            xmax=4,
            xmin=1,
            ymin=5,
            ymax=85,
            x=2cm, 
            y=0.45cm/10
    ]
            \addplot[ForestGreen, mark=o, line width=1.25pt] table [x={x}, y={Ours}] {\table};
            \addlegendentry{Ours}
            \addplot[Bittersweet, mark=pentagon,line width=1.25pt] table [x={x}, y={iCaRL}] {\table};
            \addlegendentry{iCaRL}
            \addplot[NavyBlue, mark=x,line width=1.25pt] table [x={x}, y={LwF}] {\table};
            \addlegendentry{LwF}   

        \end{axis}
        \hspace{0.5cm}
            \begin{axis}[
            at={(0.5\linewidth,0)},
            axis lines=box,
            axis line style={->},
            xlabel={Task},
            legend style={at={(0.95,0.1)},anchor=south, legend columns=5},
            x label style={at={(axis description cs:0.5,-0.1)},anchor=north},
            y label style={at={(axis description cs:0.1,0.5)},anchor=south},
            xtick={0, 1, 2, 3, 4},
            yticklabels={},
            xmax=4,
            xmin=1,
            ymin=5,
            ymax=85,
            x=2cm, 
            y=0.45cm/10,
    ]
            \addplot[ForestGreen, mark=o, line width=1.25pt] table [x={x2}, y={Ours2}] {\table};
            \addplot[Bittersweet, mark=pentagon,line width=1.25pt] table [x={x2}, y={iCaRL2}] {\table};
            \addplot[NavyBlue, mark=x,line width=1.25pt] table [x={x}, y={LwF2}] {\table};

        \end{axis}
    \end{tikzpicture}}
\end{center}
\end{subfigure}

\caption{
Comparison of the task-wise pose estimation accuracy on P3D for $4$ even tasks, where we show the thresholds \textbf{left:} $\pi/6$ and \textbf{right:} $\pi/18$.
One can observe that our method outperforms all other methods and retains high pose estimation accuracy throughout the incremental training process. One can also observe that for pose estimation, there is a stronger dependence on the difficulty of the considered classes instead of the method's ability to retain knowledge. 
}
\label{fig: peLosses}
\end{figure}
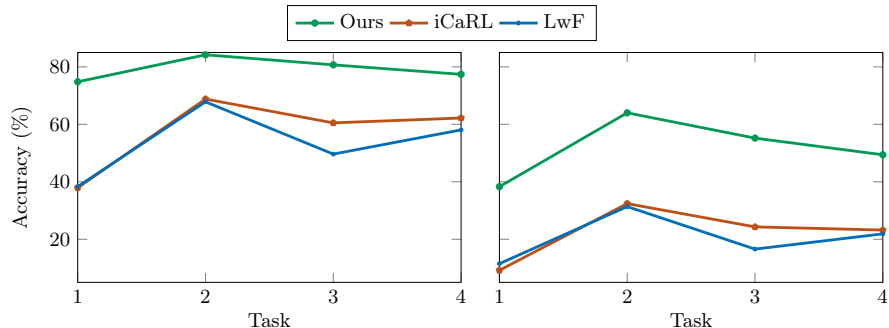

\subsection{Class Incremental Pose Estimation}
Table~\ref{table: robustness} also shows that our method significantly outperforms both ResNet-50 based methods for the task of incremental pose estimation.
As visible, the feature representation learned by the 2D pose estimation networks is much less affected by both, catastrophic forgetting and domain shifts.
Figure~\ref{fig: peLosses} shows that the performance decrease is much less severe across all tasks, where the difference in performance is much more dependent on the difficulty of the considered classes instead of the method's ability to retain knowledge.


\section{Conclusions}

In this work, we introduce incremental neural mesh models, which enable robust class-incremental learning for both, image classification and 3D pose estimation. For the first time, we present a model 
that can learn new prototypical 3D representations of object categories over time. The extensive evaluation on Pascal3D and ObjectNet3D shows that our approach outperforms all baselines even in the in-domain setting and surpasses them by a large margin in the OOD case. We also introduced the first approach for class-incremental learning of pose estimation. The results overall demonstrate the fundamental advantage of 3D object-centric representations, and we hope that this will spur a new line of research in the community.

\newpage
\section*{Acknowledgements}
We gratefully acknowledge the stimulating research environment of the GRK 2853/1 “Neuroexplicit Models of Language, Vision, and Action”, funded by the Deutsche Forschungsgemeinschaft (DFG, German Research Foundation) under project number 471607914.
Adam Kortylewski gratefully acknowledges support for his Emmy Noether Research Group, funded by the German Research Foundation (DFG) under Grant No. 468670075. Alan L. Yuille gratefully acknowledges the Army Research Laboratory award W911NF2320008 and ONR N00014-21-1-2812. 

\bibliographystyle{splncs04}
\bibliography{main}

\begin{thebibliography}{10}
\providecommand{\url}[1]{\texttt{#1}}
\providecommand{\urlprefix}{URL }
\providecommand{\doi}[1]{https://doi.org/#1}

\bibitem{Aljundii2017ExpertGate}
Aljundi, R., Chakravarty, P., Tuytelaars, T.: Expert gate: Lifelong learning with a network of experts. In: CVPR. pp. 3366--3375 (2017)

\bibitem{Aljundi2019TaskFree}
Aljundi, R., Kelchtermans, K., Tuytelaars, T.: Task-free continual learning. In: CVPR. pp. 11254--11263 (2019)

\bibitem{bai2020coke}
Bai, Y., Wang, A., Kortylewski, A., Yuille, A.: Coke: Localized contrastive learning for robust keypoint detection. Proceedings of the IEEE/CVF Winter Conference on Applications of Computer Vision  (2023)

\bibitem{Bang2021Rainbow}
Bang, J., Kim, H., Yoo, Y., Ha, J.W., Choi, J.: Rainbow memory: Continual learning with a memory of diverse samples. In: CVPR. pp. 8218--8227 (2021)

\bibitem{caron2021emerging}
Caron, M., Touvron, H., Misra, I., J\'egou, H., Mairal, J., Bojanowski, P., Joulin, A.: Emerging properties in self-supervised vision transformers. In: Proceedings of the International Conference on Computer Vision (ICCV) (2021)

\bibitem{castro2018end}
Castro, F.M., Mar{\'\i}n-Jim{\'e}nez, M.J., Guil, N., Schmid, C., Alahari, K.: End-to-end incremental learning. In: ECCV. pp. 233--248 (2018)

\bibitem{chaudhry2018riemannian}
Chaudhry, A., Dokania, P.K., Ajanthan, T., Torr, P.H.: Riemannian walk for incremental learning: Understanding forgetting and intransigence. In: ECCV. pp. 532--547 (2018)

\bibitem{Chaudhry2019AGEM}
Chaudhry, A., Ranzato, M., Rohrbach, M., Elhoseiny, M.: Efficient lifelong learning with {A-GEM}. In: ICLR (2019)

\bibitem{Chen2018Lifelong}
Chen, Z., Liu, B.: Lifelong machine learning. Synthesis Lectures on Artificial Intelligence and Machine Learning  \textbf{12}(3),  1--207 (2018)

\bibitem{cimpoi14describing}
Cimpoi, M., Maji, S., Kokkinos, I., Mohamed, S., , Vedaldi, A.: Describing textures in the wild. In: Proceedings of the {IEEE} Conf. on Computer Vision and Pattern Recognition ({CVPR}) (2014)

\bibitem{De2021ContinualSurvey}
De~Lange, M., Aljundi, R., Masana, M., Parisot, S., Jia, X., Leonardis, A., Slabaugh, G., Tuytelaars, T.: A continual learning survey: Defying forgetting in classification tasks. TPAMI  \textbf{44}(7),  3366--3385 (2021)

\bibitem{Douillard2020PODNet}
Douillard, A., Cord, M., Ollion, C., Robert, T., Valle, E.: Podnet: Pooled outputs distillation for small-tasks incremental learning. In: ECCV. pp. 86--102 (2020)

\bibitem{fecam}
Goswami, D., Liu, Y., Twardowski, B., van~de Weijer, J.: Fecam: Exploiting the heterogeneity of class distributions in exemplar-free continual learning. In: Advances in Neural Information Processing Systems. vol.~36 (2024)

\bibitem{he2016deep}
He, K., Zhang, X., Ren, S., Sun, J.: Deep residual learning for image recognition. In: CVPR. pp. 770--778 (2016)

\bibitem{hendrycks2019benchmarking}
Hendrycks, D., Dietterich, T.: Benchmarking neural network robustness to common corruptions and perturbations. In: ICLR (2019)

\bibitem{hendrycks2019augmix}
Hendrycks, D., Mu, N., Cubuk, E.D., Zoph, B., Gilmer, J., Lakshminarayanan, B.: Augmix: A simple data processing method to improve robustness and uncertainty. arXiv preprint arXiv:1912.02781  (2019)

\bibitem{Hinton2015KD}
Hinton, G., Vinyals, O., Dean, J., et~al.: Distilling the knowledge in a neural network. In: NIPS Workshops (2014)

\bibitem{Hou2019LUCIR}
Hou, S., Pan, X., Loy, C.C., Wang, Z., Lin, D.: Learning a unified classifier incrementally via rebalancing. In: CVPR. pp. 831--839 (2019)

\bibitem{iwase2021repose}
Iwase, S., Liu, X., Khirodkar, R., Yokota, R., Kitani, K.M.: Repose: Fast 6d object pose refinement via deep texture rendering. In: ICCV. pp. 3303--3312 (2021)

\bibitem{jesslen2023robust}
Jesslen, A., Zhang, G., Wang, A., Yuille, A., Kortylewski, A.: Robust 3d-aware object classification via discriminative render-and-compare. arXiv preprint arXiv:2305.14668  (2023)

\bibitem{Joseph2022Energy}
Joseph, K.J., Khan, S., Khan, F.S., Anwer, R.M., Balasubramanian, V.N.: Energy-based latent aligner for incremental learning. In: CVPR. pp. 7452--7461 (2022)

\bibitem{Kirkpatrick2017Overcoming}
Kirkpatrick, J., Pascanu, R., Rabinowitz, N., Veness, J., Desjardins, G., Rusu, A.A., Milan, K., Quan, J., Ramalho, T., Grabska-Barwinska, A., et~al.: Overcoming catastrophic forgetting in neural networks. PNAS pp. 3521--3526 (2017)

\bibitem{kortylewski2020compositionalijcv}
Kortylewski, A., Liu, Q., Wang, A., Sun, Y., Yuille, A.: Compositional convolutional neural networks: A robust and interpretable model for object recognition under occlusion. IJCV pp. 1--25 (2020)

\bibitem{kouros2022category}
Kouros, G., Shrivastava, S., Picron, C., Nagesh, S., Chakravarty, P., Tuytelaars, T.: Category-level pose retrieval with contrastive features learnt with occlusion augmentation. arXiv preprint arXiv:2208.06195  (2022)

\bibitem{krizhevsky2009learning}
Krizhevsky, A., Hinton, G., et~al.: Learning multiple layers of features from tiny images. Technical Report TR-2009  (2009)

\bibitem{li2018deepim}
Li, Y., Wang, G., Ji, X., Xiang, Y., Fox, D.: Deepim: Deep iterative matching for 6d pose estimation. In: ECCV. pp. 683--698 (2018)

\bibitem{LiH2018LwF}
Li, Z., Hoiem, D.: Learning without forgetting. TPAMI  \textbf{40}(12),  2935--2947 (2018)

\bibitem{Liu2023Online}
Liu, Y., Li, Y., Schiele, B., Sun, Q.: Online hyperparameter optimization for class-incremental learning. In: AAAI (2023)

\bibitem{Liu2024PlaceboCIL}
Liu, Y., Li, Y., Schiele, B., Sun, Q.: Wakening past concepts without past data: Class-incremental learning from online placebos. In: WACV. pp. 2226--2235 (January 2024)

\bibitem{Liu2020AANets}
Liu, Y., Schiele, B., Sun, Q.: Adaptive aggregation networks for class-incremental learning. In: CVPR. pp. 2544--2553 (2021)

\bibitem{Liu2021RMM}
Liu, Y., Schiele, B., Sun, Q.: {RMM:} reinforced memory management for class-incremental learning. In: NeurIPS. pp. 3478--3490 (2021)

\bibitem{Liu2020Mnemonics}
Liu, Y., Su, Y., Liu, A., Schiele, B., Sun, Q.: Mnemonics training: Multi-class incremental learning without forgetting. In: CVPR. pp. 12245--12254 (2020)

\bibitem{liu2021swin}
Liu, Z., Lin, Y., Cao, Y., Hu, H., Wei, Y., Zhang, Z., Lin, S., Guo, B.: Swin transformer: Hierarchical vision transformer using shifted windows. In: ICCV. pp. 10012--10022 (2021)

\bibitem{LopezPazR2017GEM}
Lopez{-}Paz, D., Ranzato, M.: Gradient episodic memory for continual learning. In: NIPS. pp. 6467--6476 (2017)

\bibitem{Luo2023CVPR}
Luo, Z., Liu, Y., Schiele, B., Sun, Q.: Class-incremental exemplar compression for class-incremental learning. In: CVPR. pp. 11371--11380. {IEEE} (2023)

\bibitem{Ma20226DNeMo}
Ma, W., Wang, A., Yuille, A.L., Kortylewski, A.: Robust category-level 6d pose estimation with coarse-to-fine rendering of neural features. In: ECCV. pp. 492--508 (2022)

\bibitem{McCloskey1989Catastrophic}
McCloskey, M., Cohen, N.J.: Catastrophic interference in connectionist networks: The sequential learning problem. In: Psychology of Learning and Motivation, vol.~24, pp. 109--165. Elsevier (1989)

\bibitem{McRae1993Catastrophic}
McRae, K., Hetherington, P.: Catastrophic interference is eliminated in pre-trained networks. In: CogSci (1993)

\bibitem{imagecorruptions}
Michaelis, C., Mitzkus, B., Geirhos, R., Rusak, E., Bringmann, O., Ecker, A.S., Bethge, M., Brendel, W.: Benchmarking robustness in object detection: Autonomous driving when winter is coming. arXiv preprint arXiv:1907.07484  (2019)

\bibitem{Mousavian20173D}
Mousavian, A., Anguelov, D., Flynn, J., Kosecka, J.: 3d bounding box estimation using deep learning and geometry. In: CVPR. pp. 7074--7082 (2017)

\bibitem{papyan2020prevalence}
Papyan, V., Han, X., Donoho, D.L.: Prevalence of neural collapse during the terminal phase of deep learning training. Proceedings of the National Academy of Sciences  \textbf{117}(40),  24652--24663 (2020)

\bibitem{fetril}
Petit, G., Popescu, A., Schindler, H., Picard, D., Delezoide, B.: Fetril: Feature translation for exemplar-free class-incremental learning. In: CVPR (2023)

\bibitem{Pourkeshavarzi2022Looking}
PourKeshavarzi, M., Zhao, G., Sabokrou, M.: Looking back on learned experiences for class/task incremental learning. In: ICLR (2022)

\bibitem{Prabhu2020GDumb}
Prabhu, A., Torr, P.H., Dokania, P.K.: {GD}umb: A simple approach that questions our progress in continual learning. In: ECCV. pp. 524--540 (2020)

\bibitem{ravi2020pytorch3d}
Ravi, N., Reizenstein, J., Novotny, D., Gordon, T., Lo, W.Y., Johnson, J., Gkioxari, G.: Accelerating 3d deep learning with pytorch3d. arXiv:2007.08501  (2020)

\bibitem{Rebuffi2017iCaRL}
Rebuffi, S.A., Kolesnikov, A., Sperl, G., Lampert, C.H.: {iCaRL}: Incremental classifier and representation learning. In: CVPR. pp. 5533--5542 (2017)

\bibitem{russakovsky2015imagenet}
Russakovsky, O., Deng, J., Su, H., Krause, J., Satheesh, S., Ma, S., Huang, Z., Karpathy, A., Khosla, A., Bernstein, M., et~al.: Imagenet large scale visual recognition challenge. International journal of computer vision  \textbf{115},  211--252 (2015)

\bibitem{Shin2017GenerativeReplay}
Shin, H., Lee, J.K., Kim, J., Kim, J.: Continual learning with deep generative replay. In: NeurIPS. pp. 2990--2999 (2017)

\bibitem{Simon2021Learning}
Simon, C., Koniusz, P., Harandi, M.: On learning the geodesic path for incremental learning. In: CVPR. pp. 1591--1600 (2021)

\bibitem{Tao2020Topology}
Tao, X., Chang, X., Hong, X., Wei, X., Gong, Y.: Topology-preserving class-incremental learning. In: ECCV. pp. 254--270 (2020)

\bibitem{Tulsiani2015Viewpoints}
Tulsiani, S., Malik, J.: Viewpoints and keypoints. In: CVPR (June 2015)

\bibitem{vaswani2017attention}
Vaswani, A., Shazeer, N., Parmar, N., Uszkoreit, J., Jones, L., Gomez, A.N., Kaiser, {\L}., Polosukhin, I.: Attention is all you need. NeurIPS  \textbf{30} (2017)

\bibitem{wang_nemo}
Wang, A., Kortylewski, A., Yuille, A.: {NeMo}: Neural mesh models of contrastive features for robust 3d pose estimation. ICLR  (2021)

\bibitem{wang2024neural}
Wang, A., Ma, W., Yuille, A., Kortylewski, A.: Neural textured deformable meshes for robust analysis-by-synthesis. In: WACV. pp. 3108--3117 (2024)

\bibitem{wang2021neural}
Wang, A., Mei, S., Yuille, A.L., Kortylewski, A.: Neural view synthesis and matching for semi-supervised few-shot learning of 3d pose. NeurIPS  \textbf{34},  7207--7219 (2021)

\bibitem{wang2020robust}
Wang, A., Sun, Y., Kortylewski, A., Yuille, A.L.: Robust object detection under occlusion with context-aware compositionalnets. In: CVPR. pp. 12645--12654 (2020)

\bibitem{wang2022voge}
Wang, A., Wang, P., Sun, J., Kortylewski, A., Yuille, A.: Voge: a differentiable volume renderer using gaussian ellipsoids for analysis-by-synthesis. In: ICLR (2022)

\bibitem{Wang2022FOSTER}
Wang, F.Y., Zhou, D.W., Ye, H.J., Zhan, D.C.: Foster: Feature boosting and compression for class-incremental learning. In: ECCV (2022)

\bibitem{wu2018memory}
Wu, C., Herranz, L., Liu, X., Van De~Weijer, J., Raducanu, B., et~al.: Memory replay gans: Learning to generate new categories without forgetting. NeurIPS  \textbf{31} (2018)

\bibitem{wu2019large}
Wu, Y., Chen, Y., Wang, L., Ye, Y., Liu, Z., Guo, Y., Fu, Y.: Large scale incremental learning. In: CVPR. pp. 374--382 (2019)

\bibitem{xiang2016objectnet3d}
Xiang, Y., Kim, W., Chen, W., Ji, J., Choy, C., Su, H., Mottaghi, R., Guibas, L., Savarese, S.: Objectnet3d: A large scale database for 3d object recognition. In: ECCV (2016)

\bibitem{xiang2014beyond}
Xiang, Y., Mottaghi, R., Savarese, S.: Beyond pascal: A benchmark for 3d object detection in the wild. In: WACV. pp. 75--82. IEEE (2014)

\bibitem{xiang_wacv14}
Xiang, Y., Mottaghi, R., Savarese, S.: Beyond pascal: A benchmark for 3d object detection in the wild. In: WACV (2014)

\bibitem{yan2021dynamically}
Yan, S., Xie, J., He, X.: Der: Dynamically expandable representation for class incremental learning. In: CVPR. pp. 3014--3023 (2021)

\bibitem{zhao2022ood}
Zhao, B., Yu, S., Ma, W., Yu, M., Mei, S., Wang, A., He, J., Yuille, A., Kortylewski, A.: Ood-cv: A benchmark for robustness to individual nuisances in real-world out-of-distribution shifts. In: ECCV (2022)

\bibitem{Zhou2023PyCIL}
Zhou, D., Wang, F., Ye, H., Zhan, D.: Pycil: a python toolbox for class-incremental learning. Sci. China Inf. Sci.  \textbf{66}(9) (2023)

\bibitem{zhou2018starmap}
Zhou, X., Karpur, A., Luo, L., Huang, Q.: Starmap for category-agnostic keypoint and viewpoint estimation. In: ECCV. pp. 318--334 (2018)

\end{thebibliography}
\newpage
\appendix

\section*{Supplementary Material for iNeMo: Incremental Neural Mesh Models for Robust Class-Incremental Learning}

In the following, we provide further details and ablation studies for our paper. 
In the first section we define the conventions. We then provide the non-incremental performance of both considered representations (R50 and NeMo) as a reference. Afterwards, we show the advantage of considering uncertainty for the classification in Section~\ref{AppInference}  and then give a conclusive ablation study over all  components of our method in Section~\ref{AppAblation}.
Since NeMo is trained with additional pose labels that were not available to baselines, we provide an additional study in Section~\ref{AppPEClassifier} where we show that pose labels do not improve the baselines. 
Finally, we conclude with additional details  on the background model and pose estimation, as well as all the  considered hyperparameters in our method.

\section{Conventions}

In the tables of the main paper, we followed previous work~\cite{Rebuffi2017iCaRL,LiH2018LwF,Liu2020Mnemonics} and reported the average of the testing accuracies over all tasks $\mathcal{T}^i$ with $\Phi_i$, which we denoted as $acc(1\,{\colon}i)$ .
In the supplemental material, we deviate from this setting and \textbf{report the final accuracy with} $\Phi_{N_{\text{task}}}$ \textbf{on the whole test dataset after integrating all tasks},  as it determines the final performance loss that is usually most significant. 
We denote the final accuracy on all seen classes after training on the final task $\mathcal{T}^{N_{\text{task}}}$ as $\overline{acc}(1\,{\colon}N_{\text{task}})$.


\section{Non-Incremental Upper Bounds}

To determine an upper bound for the performance of ResNet50 and NeMo approaches, 
we train on all classes jointly and provide the results in
Table~\ref{table: Upperbound}. 
While both approaches are able to achieve similar performance for classification on P3D, NeMo significantly outperforms the RestNet50 on O3D. We suspect that the reason for this is that O3D contains a large number of occluded and truncated objects. NeMo generally outperforms the ResNet50 for pose estimation implemented following~\cite{zhou2018starmap}. 



{
\setlength{\tabcolsep}{1.8mm}{
\begin{table*}[tp]
  \small
  \centering
    \caption{
  We trained the RestNet50 and NeMo approaches jointly on all classes to determine an upper bound for their performance. All networks were initialized with weights from DINOv1~\cite{caron2021emerging}, which itself was trained in an unsupervised fashion.
  For joint training of NeMo, we follow the training protocol of Jesslen \textit{et al.}~\cite{jesslen2023robust} with the exception of using pre-trained weights as mentioned. 
  }
  \begin{tabu}{llccccc}
  \toprule
  Metric & & Type & Repr.&  \emph{P3D} &  \emph{O3D}\\
  \midrule
    Classification & & Joint &R50&98.32&76.23\\
    $\overline{acc}(1\,{\colon}N_{\text{task}})$ in \% $\uparrow$& & Joint& NeMo   &99.27&85.28 \\
    \bottomrule
     & &  & &   & \\
     \toprule
    Metric & & Type & Repr.&  \emph{P3D} & \\
    \midrule
    Pose $\pi/6$ & & Joint &R50&74.6& \\
    $\overline{acc}(1\,{\colon}N_{\text{task}})$ in \% $\uparrow$& & Joint& NeMo   &87.25& \\
    \midrule
    Pose $\pi/18$ & & Joint &R50&36.5& \\
    $\overline{acc}(1\,{\colon}N_{\text{task}})$ in \% $\uparrow$& & Joint& NeMo   &65.81& \\
  \bottomrule
    \end{tabu}
  \label{table: Upperbound}

\end{table*}
}
}

{
\setlength{\tabcolsep}{1.8mm}{
\begin{table*}[tp]
  \small
  \centering
    \caption{
  We compare our inference approach to the one proposed by Jesslen \textit{et al.}~\cite{jesslen2023robust}.
  When training jointly, the performance is nearly identical.
  However, when training incrementally, disentangling visually similar features becomes more challenging and our proposed strategy significantly improves the result. 
  }
  \begin{tabu}{llcccccc}
  \toprule
  \multirow{2.5}{*}{Metric} & & \multirow{2.5}{*}{Type} & \multirow{2.5}{*}{Inference}&&  \emph{P3D} &  \emph{O3D}\\
  \cmidrule{6-6} \cmidrule{7-7} 
  \rowfont{\tiny}& &    &  & & $B0+3$ & $B0+20$  \\
  \midrule
    & & Joint & \cite{jesslen2023robust} &&99.28& 85.28 \\
    Classification
    & & Joint & Ours &&99.27&85.28\\
    \cmidrule{3-7}
    \multirow{1}{*}{$\overline{acc}(1\,{\colon}N_{\text{task}})$ in \% $\uparrow$
    }& & Incremental& \cite{jesslen2023robust}   &&95.06&75.8 \\
    & & Incremental& Ours    &&96.41&80.17\\
  \bottomrule
    \end{tabu}

  \label{table: Inference}

\end{table*}
}
}

\section{Considering Uncertainty in Classification}
\label{AppInference}

We proposed an extension to the classification strategy introduced by Jesslen \textit{et al.}~\cite{jesslen2023robust} in Equation 14 of the main paper which was motivated by the observation that classes sharing visually similar features were confused more often when training the model in an incremental setting. We believe that when training on all classes jointly, the contrastive loss between all features of different classes is sufficient to ensure that all parts of different objects have distinct feature representations. However, such disentanglement is significantly more challenging in an incremental setting. 
The results from Table~\ref{table: Inference} show that our proposed strategy to exclude pixels that ambiguously relate to meshes of multiple possible classes (i.e. uncertain pixels) brings a significant improvement.




{

\setlength{\tabcolsep}{1.8mm}{
\begin{table*}[tp]
  \small
  \centering
    \caption{
  \textbf{Top:}
  We provide an ablation study for the 2D ResNet50 and NeMo with the traditional class-incremental techniques LwF and iCaRL. As visible, traditional techniques work less well on NeMo. \textbf{Bottom:} We provide an ablation study of our model components and show that all of our additions increase the performance. 
  We indicate the used exemplar selection strategy in the column "Replay", where H denotes the herding strategy~\cite{Rebuffi2017iCaRL} and PA our pose-aware exemplar selection strategy.
  Note that we used our improved inference method from Section~\ref{AppInference} for all methods.}
  \scalebox{0.93}{
  \begin{tabu}{lcc|c|c|c|c|cccc}
  \toprule
  \multirow{2.5}{*}{Metric} & \multirow{2.5}{*}{Method} & \multirow{2.5}{*}{Repr.} & \multirow{2.5}{*}{\scriptsize Replay} &\multirow{2.5}{*}{\scriptsize Init} &\multirow{2.5}{*}{\scriptsize Pos.}&\multirow{2.5}{*}{\scriptsize KD}& & \emph{P3D} &  \emph{O3D}\\
  \cmidrule{9-9} \cmidrule{10-10} 
  \rowfont{\tiny}&     &  &&&&& & $B0+3$ & $B0+20$  \\
  \midrule
     &  Finetune & NeMo & - & & & &  &17.47&  17.81\\
    &  LwF & R50 &- &&& \checkmark &  &83.75&  56.44\\
    & LwF & NeMo &- &&& \checkmark &  &17.45&  17.72\\
    &  iCaRL & R50 &H &&&  \checkmark &  &91.79&61.75\\
    Classification
    &  iCaRL & NeMo &H &&&  \checkmark &  &93.72&68.87\\
    \cmidrule{2-10}
    \multirow{1}{*}{$\overline{acc}(1\,{\colon}N_{\text{task}})$ in \% $\uparrow$
    }&  Ours& NeMo &H&&&  &   &93.60&69.01 \\
    & Ours& NeMo &PA&&&  &   &94.70&70.32 \\
    &  Ours& NeMo &PA& \checkmark& &   &   &94.78&71.67\\
    &  Ours& NeMo &PA& \checkmark&\checkmark &   &   &94.98&72.09\\
    &  Ours & NeMo &PA&\checkmark&\checkmark& \checkmark & &96.41&80.17\\
  \bottomrule
    \end{tabu}}

  \label{table: Ablation}

\end{table*}
}
}
\section{Ablation}
\label{AppAblation}

In the main paper, we have shown that our novel class-incremental learning strategy with neural meshes significantly outperforms the 2D baselines. In the following, we provide an analysis of how much the individual parts of our model contribute to this result. 


Table~\ref{table: Ablation} shows the contribution of each of our model components.
We start with the most naive extension of NeMo to the class-incremental setting:
in each task, we initialize the required number of meshes and fine-tune the feature extractor on each new task dataset. As expected, this leads to bad results. 
Next, we extend the models by the traditional distillation~\cite{LiH2018LwF} (LwF) and herding exemplar~\cite{Rebuffi2017iCaRL} (iCaRL) strategies. The latter brings significant improvements. This shows that overall exemplar replay is necessary to retain knowledge while training Neural Mesh Models incrementally. We also compare applying LwF and iCaRL to either the 2D ResNet50 or NeMo and find that those strategies in most cases work better for the 2D setting, hence not being simply transferable to NeMo. 

We finally demonstrate that our additions to maintain a structured latent space provide the best results by introducing the latent space initialization, positional regularization, and adding knowledge distillation. The results indicate that knowledge distillation has little effect (row 9), while adding the pose-aware replay (row 5 to row 6) has the largest impact on the result. This shows that the pose-aware  exemplar selection strategy is critical and all other additions further improve the performance. 




\section{Training with less Replay Memory}

Memory replay is essential when training iNeMo, as it allows updating neural meshes from previous tasks. 
However, storing too many samples per class in memory can become quite expensive and as such it is crucial for methods to be effective in utilizing replay with fewer samples.
We show in Table~\ref{table: Replay} that iNeMo can adapt to lower memory sizes, but is optimal for the chosen 20 exemplars.

{
\setlength{\tabcolsep}{5.2mm}{
\begin{table}[tp]
  \small
  \centering
    \caption{
  Final task accuracy on Pascal3D with decreasing number of exemplars per class. Even with few exemplars, iNeMo retains good accuracy.
  }
  \begin{tabu}{llc}
  \toprule
  \multirow{2.5}{*}{Metric} & \multirow{2.5}{*}{Exemplars}  & \emph{P3D} \\
  \cmidrule{3-3} 
  \rowfont{\tiny} &   & $B0+3$ \\
    \midrule
     Classification & 20 & 96.41 \\
     $\overline{acc}(1\,{\colon}N_{\text{task}})$ in \% $\uparrow$ & 10 & 93.36 \\
     & \hphantom{0}5 & 82.59\\
  \bottomrule
\end{tabu}

  \label{table: Replay}
\end{table}
}
}

\section{Enhancing 2D Classifiers with Pose Annotations}
\label{AppPEClassifier}

Neural Mesh Models leverage meshes to host 3D consistent features and consequently, their training requires camera pose annotations. However, such pose annotations were not used in the 2D baselines, which could in principle give the Neural Mesh Models an advantage. To this end, we evaluate if using the pose annotation could improve the results of the 2D baselines. We extend the ResNet50 model with a second classifier head to predict the pose following~\cite{zhou2018starmap} and use the following combined loss: 
\begin{equation}
    \mathcal{L}_{pe-cl} = \mathcal{L}_{iCaRL} + \lambda_{pe}\mathcal{L}_{Starmap}.
\end{equation}
We then train the models in a class-incremental fashion with the iCaRL~\cite{Rebuffi2017iCaRL} strategy. 
The results are provided in 
Table~\ref{table: pe-cl} and show that the additional pose supervision introduced in this way does not help to improve the classification accuracy.
When increasing the weight of the pose loss $\lambda_{pe}$, the performance consistently decreases with the best performing model being the default iCaRL network with $\lambda_{pe}=0$.
{
\setlength{\tabcolsep}{1.8mm}{
\begin{table*}[tp]
  \small
  \centering
    \caption{
    Adding an additional pose estimation head and providing additional supervision does not lead to better representation learning.
    It is not obvious how conventional classifiers could leverage the additional camera pose annotation.
  }
  \begin{tabu}{llccc}
  \toprule
\multirow{2.5}{*}{Metric} & \multirow{2.5}{*}{$\lambda_{pe}$}  & \emph{P3D}\\
 \cmidrule{3-3} 
  \rowfont{\tiny} &   & $B0+3$ \\
  \midrule
        & 0.00   &91.79\\
    Classification  & 0.33 &91.42\\
    $\overline{acc}(1\,{\colon}N_{\text{task}})$ in \% $\uparrow$ & 0.66   &90.56\\
    & 1.00   &89.86\\
    \bottomrule
    \end{tabu}

  \label{table: pe-cl}

\end{table*}
}
}

\section{Additional Implementational Details}

In this section, we provide the full implementation details about our method.

\subsubsection{Data Preparation.}

NeMo~\cite{wang_nemo} was originally proposed for 3D pose estimation, which means that the degrees of freedom to the camera pose are azimuth, elevation, and roll angle. This implies that the objects are scaled accordingly and centered in the images. We follow this procedure and use the publicly available code of NeMo.
To make the sizes of all input images consistent, we further pad all images to the size of $640\times 800$, where we fill the padded region with random textures from the Describable Textures Dataset~\cite{cimpoi14describing}.

\subsubsection{Obtaining the 3D Cuboid Mesh} is possible, since P3D~\cite{xiang2014beyond} and O3D~\cite{xiang2016objectnet3d} provide a selection of 3D CAD models for each object category.
For our 3D cuboid mesh, we consider the average bounding box of those models. We then sample vertices uniformly on its surface, leading to roughly 1,100 vertices per mesh.

\subsubsection{Annotations} at training time are computed with PyTorch3D's~\cite{ravi2020pytorch3d} mesh rasterizer.
Concretely, we render the neural meshes with ground-truth camera poses to compute their projection and binary object masks.
Additionally, we compute the projected coordinates $\pi(k)$ of each vertex $V^k$ and its binary visibility $o^k$.  
Given the class label, we render the corresponding mesh at $\frac{1}{8}$ of the original image resolution (same size as the output of the feature extractor $\Phi$).
At each pixel, we determine vertex visibility by considering the closest face using the returned z-buffer. 
To parameterize the rasterizer, we use a relatively simple camera model with a focal length of $1$.
As there is no viewport specified for neither the P3D or OOD-CV~\cite{zhao2022ood} dataset, we follow previous work~\cite{jesslen2023robust, wang_nemo, wang2022voge} and use a viewport of $3000 / 8$.
For the O3D~\cite{xiang2016objectnet3d} dataset we use their specified viewport of $2000/8$.

\section{Pose Estimation}

For the pose estimation, we follow previous work~\cite{wang_nemo, jesslen2023robust}.
For completeness, we also provide a brief explanation here on how one can estimate the 3D object pose of an object of class $c$, given the trained feature extractor $\Phi$ and the neural mesh $\mathfrak{N}_c$.

\subsubsection{3D Pose Estimation.}
During inference, we do not have access to the camera pose and corresponding perspective transformation.
Since the camera intrinsics and distance to the object are assumed to be known a-priori, we need to optimize for the unknown camera pose $Q^{\text{pred}}$.
We define the foreground $\mathcal{F}$ in the same way as we did for the classification in Section 4.4 of the main part of the paper.
However, in addition to pixels that have been recognized as background, we also remove pixel positions that fall outside the projection of the cuboid, leading to $\mathcal{F}_{\pi^{\text{pred}}}=\mathcal{F}\cap \{f_{\pi^{\text{pred}}(k)}| \,\forall\, V_k \in \mathcal{V}_c \mathrm{\, \wedge \,} o_c^k=1 \}$.

\subsubsection{Finding $Q^{\text{pred}}$} is done via a render-and-compare approach.
We do so by initializing a rough estimate and optimizing iteratively.
Given the current camera pose and its incurred perspective transform $\pi^{\text{pred}}$, we maximize the current object likelihood:
\begin{equation}
   \max_{Q^{\text{pred}}} \prod_{f_{\pi^{\text{pred}}(k)}\in \mathcal{F}_{\pi^{\text{pred}}}} P(f_{\pi^{\text{pred}}(k)}|\theta^k_c).
\end{equation}
By considering the vMF distribution, we optimize the initial camera pose using PyTorch3D's~\cite{ravi2020pytorch3d} differentiable rasterizer by minimizing the negative log likelihood:
\begin{equation}
    \mathcal{L}_{RC}(Q^{\text{pred}}) = \sum_{f_{\pi^{\text{pred}}(k)}\in \mathcal{F}_{\pi^{\text{pred}}}} -f_{\pi^{\text{pred}}(k)}^\top \cdot \theta^k_c.
\end{equation}

\subsubsection{Efficient Pose Estimation via Template Matching.}
The convergence of the above process is highly reliant on the provided initial pose, making it prohibitively slow in a worst case scenario.
Wang \textit{et al.}~\cite{wang2021neural} proposed to speed it up by pre-rendering all neural meshes from $144$ distinct viewing angles.
Before the render-and-compare process, the output of the feature extractor is compared to each of these pre-rendered maps and the camera pose $Q^{\text{pred}}$ is initialized with the pose that maximized the object likelihood.
This simple procedure is remarkably effective, giving a speed-up of approximately $80\%$~\cite{jesslen2023robust} over the original approach~\cite{wang_nemo}.

\section{Modelling the Background}

For both classification and pose estimation, we leverage a set of features $\mathcal{B}$.
This approach of disentangling foreground and background features into separate sets was introduced by Bai \textit{et al.}~\cite{bai2020coke}.
Although we do not have a combined foreground set (but rather separate, 3D consistent meshes), we adopt their handling of the background model.

\subsubsection{Learning the Background Model.}

We maintain a set $\mathcal{B}$ of $N_{\text{bg}}$ features that are sampled from positions in the feature map that fall outside the cuboid projection.
From each sample in a training batch of size $b$, we sample $N_{\text{bgupdate}}$ new background features.
Consequently, we need to remove $b\cdot N_{\text{bgupdate}}$ from $\mathcal{B}$ to avoid going over the allocated memory limit.
Replacement is done, by maintaining a counter for each background feature, that indicates how many update steps it has been alive in $\mathcal{B}$ and prioritizing the oldest features for removal.  

\subsubsection{Balancing the Background Model.}

Ideally, the background should contain features from a wide variety of background options (\ie water from boats, sky from airplanes, urban scenes from cars, ...).
However, sampling background features from the current task dataset only means that $\mathcal{B}$ would be heavily biased towards background features from the currently considered classes. Therefore, we balance $\mathcal{B}$ after each training phase by sampling background features from the exemplar memory $\mathcal{E}^{1:i}$, which was constructed evenly from all classes and viewing angles.

\section{Hyperparameter Collection}

As there are quite a few hyperparameters involved in our method, we include this brief section that notes down all parameters for our final model.

{
\setlength{\tabcolsep}{1.8mm}{
\begin{table*}[ht]
  \small
  \centering
  \begin{tabu}{cccccc}
  \toprule
  Opt. & LR & $(\beta_1,\beta_2)$ & Task-Epoch & Batch Size & Weight Decay \\
  \midrule
  Adam & 1e-5  & (0.9,0.999) & 50 & 16 & 1e-4\\
  \bottomrule
    \end{tabu}
  \cotronlvsapce
  \caption{Optimization Parameters.}
\end{table*}
}
}
\vspace{-1.5cm}
{
\setlength{\tabcolsep}{1.8mm}{
\begin{table*}[ht]
  \small
  \centering

  \begin{tabu}{ccccc}
  \toprule
  $\kappa_1$ & $\kappa_2$ & $\kappa_3$ & $\lambda_{etf}$ & $\lambda_{kd}$ \\
  \midrule
  1/0.07 & 1  & 0.5 & 0.1 & 10.0\\
  \bottomrule
    \end{tabu}
    \caption{Loss Weighting.}
\end{table*}
}
}
\vspace{-1.5cm}
{
\setlength{\tabcolsep}{1.8mm}{
\begin{table*}[ht]
  \small
  \centering
  \begin{tabu}{ccccc}
  \toprule
  d & $\eta$ & $N_{\text{bg}}$ & $N_{\text{bgupdate}}$ & R \\
  \midrule
  128 & 0.9  & 2560 & 5 & 48\\
  \bottomrule
    \end{tabu}
    \caption{Mesh- and Background-related Parameters.}
\end{table*}
}
}

\vspace{-1.5cm}

{
\setlength{\tabcolsep}{1.8mm}{
\begin{table*}[ht]

  \small
  \centering

  \begin{tabu}{cccc}
  \toprule
  Opt. & LR & $(\beta_1,\beta_2)$ & Epochs \\
  \midrule
  Adam & 5e-2  & (0.4,0.6) & 30 \\
  \bottomrule
    \end{tabu}
    \caption{Pose Estimation Parameters.}
\end{table*}
}
}

\end{document}